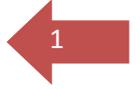

# Prépublication

Peter Stockinger

# Du modèle conceptuel des données à sa mise en scène multimodale.

## Réflexions sémiotiques sur le *design de l'information.*

Équipe « Pluralité des langues et des identités » (PLIDAM)
Institut national des langues et civilisations orientales (INALCO)

**Lambach - Paris 2018**

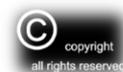

# Table des matières

## Introduction

Dans cet article, nous nous proposons de problématiser la notion du design de l'information en distinguant plus particulièrement entre :

1. une *pratique du design du sens* d'un domaine de référence ou bien d'un corpus de données textuelles et
2. une *pratique du design de l'expression, de la mise en scène visuelle* (lato sensu) du sens, de la valeur d'une donnée textuelle (cette seconde pratique correspond en gros à la définition usuelle de la notion « design de l'information », comprise comme un ensemble de techniques de représentations graphiques de la valeur d'un ensemble de données).

Dans notre discussion, nous allons nous concentrer sur un domaine d'expertise que nous pratiquons depuis une quinzaine d'années, celui de la constitution et de l'enrichissement de *corpus de données textuelles* (notamment *audiovisuelles*) et de l'*éditorialisation* de tels corpus sous forme de multiples « mises en récit » dans des formats visuels de communication les plus divers.

Nous discuterons rapidement les trois points suivants :

1. La notion du design de l'information.
2. Le design au sens de *modèle conceptuel/ modélisation conceptuelle* d'un domaine/d'un corpus de données textuelles.
3. Le design au sens de l'expression et de la mise en scène multimodale (et pas seulement visuelle) des métadonnées et des données textuelles.

Cet article a fait l'objet d'une communication que nous avons donnée dans le cadre de la Journée « Sémiotique du design d'information » organisée par Anne Beyaert-Geslin de l'université Bordeaux Montaigne (laboratoire MICA) le 21 juin 2018 à Bordeaux.

## 1) La notion « design de l'information[1] »

Une des définitions habituelles de la notion *design de l'information* est celle d'être une technique de l'expression ou de la mise en scène (visuelle : graphique…) de l'information qui, aujourd'hui, recourt aux logiciels et aux outils numériques de visualisation (services web, API …).

Le design de l'information possède déjà une longue tradition. Mentionnons ici seulement son usage dans ce qu'on appelle en anglais le *technical writing*. Le *technical writing* est l'art de structurer, de visualiser, de publier et d'actualiser des contenus souvent très spécialisés dans des domaines et des secteurs d'activités technologiques et scientifiques de pointe tels que la physique des fluides, l'aérospatiale ou l'ingénierie informatique. Un des enjeux centraux du *technical writing* est, bien sûr, la visualisation de procédés et d'instruments technologiques complexes tels que, par exemple, la structure et le fonctionnement d'un moteur d'avion ou d'un microordinateur. Les visuels sont souvent des dessins réalisés, aujourd'hui, à l'aide de logiciels spécialisés.

---

[1] Entre les deux désignations « design d'information » (traduction de l'anglais « information design ») et design de l'information », nous optons pour la deuxième qui semble avoir une signification plus large en incluant – outres les méthodes graphiques - « l'ensemble des méthodes de mise en forme de l'information au sens large » (Vollaire 1997).

Cela étant, la visualisation de processus, procédés, artefacts, etc. ne représente qu'un aspect dans un projet de *technical writing*. Un tel projet – comme tout autre projet - connaît plusieurs phases parmi lesquelles on trouve la *spécification du projet* lui-même (le « plan du projet »), l'élaboration d'une *structure adaptée du document* à réaliser, la *rédaction progressive* du contenu avec plus particulièrement la *visualisation*, l'*évaluation* et la *correction* du contenu rédigé et, enfin, sa *publication*. On voit bien que la visualisation des données ou d'information n'est pas une activité indépendante, autonome. Elle prend obligatoirement place dans un *projet de communication éditoriale* qui poursuit des objectifs précis, qui s'adresse à une certaine catégorie de destinataires et qui doit satisfaire à un ensemble d'usages circonscrits. Autrement dit, la visualisation de données (d'information, de connaissance, etc.) est une activité fonctionnelle et ne peut être évaluée que par rapport à sa place et à sa fonction dans un projet donné de communication.

En considérant d'une manière générale la notion du « design de l'information », on voit à l'œuvre la distinction essentielle entre :

1. le *contenu*, le *sens*, la *valeur* (donc : l'*univers sémantique*) d'un ensemble ou d'un corpus de données et
2. l'*expression* du contenu, du sens ou de la valeur (c'est-à-dire de *l'univers sémantique*) de cet ensemble de données ou du corpus considéré.

En d'autres termes, on retrouve ici la distinction structuraliste entre *signifié* et *signifiant* du signe qu'on pourra, à notre avis, parfaitement utiliser pour traiter visuellement les données, les données considérées comme des signes, c'est-à-dire comme des entités signifiantes possédant un sens pour un acteur donné.

Mais, le contenu d'une donnée, son sens, dans la définition habituelle du design de l'information, ne semble pas être problématisé. Il constitue une donnée sur laquelle on ne s'interroge pas ; il constitue une évidence donnée a priori. En revanche, ce que la notion du « design de l'information » problématise, c'est l'*expression*, la *mise en scène visuelle* du sens donné à priori en vue d'orienter, d'influencer, de rendre plus compréhensible sa perception et sa compréhension par un public donné.

Prenons un exemple très simple. Le fait qu'une population ayant répondu à un sondage donné soit composée de tant d'hommes et de femmes, de tant de personnes appartenant à une tranche d'âge, de tant de personnes représentant une certaine CSP, ce fait-là – cette *information* – est accepté tel quel, est accepté comme une évidence et le seul problème est celui de sa représentation sous forme d'un histogramme, d'un diagramme en bâtons, d'un graphique en rayons de soleil, etc.

La notion "design" comprise comme une *activité* possède cependant un sens similaire à celui de la *modélisation*, de la *spécification de modèles*, le modèle étant compris au sens d'une « image mentale », d'une « vision mentale » d'un objet visé, d'un objet *intentionnel*. En reconsidérant donc la compréhension habituelle de la notion « design de l'information », il me semble opportun de distinguer entre deux types de design :

1. Le *design du contenu* : c'est la spécification, la définition d'une *vue sémantique* sur un corpus de données. Pour reprendre l'exemple simple de l'*échantillon* (caractéristique, aléatoire) d'un sondage : le modèle conceptuel qui préside, qui motive la sélection, la composition de cet échantillon, c'est le *point de vue*, la *vision mentale* selon lequel nous appréhendons la population en général dont nous souhaitons connaître un certain comportement à travers la technique du sondage. De ce point de vue, cette vision s'exprime sous forme d'un ensemble d'hypothèses ou de certitudes sur le comportement de la population, ses caractéristiques sociales, son niveau et son style de vie, et ainsi de suite. Or, il s'agit bien ici d'un problème du *design d'un contenu*, de la *sémantique* d'un certain type de données (dans notre cas de la *sémantique* de la population à étudier) si on entend par *design* quelque chose qui est équivalent à l'activité de *modélisation* ainsi qu'au résultat de cette activité (c'est-à-dire au modèle, au schéma, au script, au scénario). Le *design du contenu* possède ses problèmes particuliers qui relèvent de la *description sémantique*.

2. Le *design de l'expression*, de la *mise en scène* visuelle (voire, d'une manière plus générale, polysensorielle) des données collectées, traitées, décrites ou analysées selon un point de vue (une « vision ») adopté, en référence à un modèle conceptuel. Cet aspect du design de l'information renvoie à la problématique de l'élaboration et de l'usage de langages visuels.

Dans ce qui suit, nous nous concentrerons sur le design compris comme une pratique de la *définition-spécification du sens* d'un domaine de connaissance circonscrit.

## 2) Le design du sens d'un domaine de connaissance

On utilise différentes expressions pour désigner les techniques et activités de la *définition-spécification du sens* d'un domaine de connaissance circonscrit : *cartographie sémantique, cognitive* ou encore *mentale* ; *analyse conceptuelle* ; *cognitique* ou *analyse des connaissances*, etc.

En général ces approches se réfèrent et utilisent des connaissances et savoir-faire provenant de plusieurs disciplines : de la sémantique (linguistique, cognitive, conceptuelle…), de la psychologie cognitive, de l'anthropologie cognitive, de la sémiotique structurale ou peircienne et, bien sûr, de l'informatique appliquée ainsi que, depuis plus récemment, desdites sciences des données et du web sémantique. Elles poursuivent souvent des objectifs définis dans un cadre de recherche-développement, de recherche appliquée ou industrielle.

Parmi les « outils » méthodologiques les plus centraux, les plus souvent utilisés, on trouve les *réseaux sémantiques* ou *conceptuels*. Un réseau sémantique ou conceptuel est une *configuration* qui *qualifie* (définit et décrit) – selon un *certain point de vue* et étant donné un *contexte d'usage* ou d'*application* – le *sens* d'un *objet donné*. La configuration, elle, est généralement représentée sous forme d'un *graphe*, autrement dit sous forme d'un ensemble de *sommets* (de « points ») et d'*arêtes* (de *chemins* entre *points*).

Cet « outil » a fait son apparition dans les années 50/60 et s'est popularisé fin 60 grâce à une série de travaux de recherche d'Alan M. Colins et M. Ross Quillian en psychologie cognitive consacrés à la mémoire sémantique. Dans les années 1970 et 80, des propositions similaires ont été avancées par des chercheurs en intelligence artificielle, en traitement automatique du langage et en sciences cognitives. On peut citer, par exemple, la *frame theory* (la théorie des cadres conceptuels ou cognitifs) de M. Minsky, la *théorie des scripts* (ou des scénarios) élaborée par Roger Schank et Robert P. Abelson ou encore les *réseaux contractuels* développés et utilisés dans le cadre de l'intelligence artificielle distribuée, des systèmes multiagents, etc. Aujourd'hui, les réseaux sémantiques ou conceptuels sont largement utilisés dans le cadre des langages ontologiques du web sémantique (Lalande et al, 2015 ; Beloued et al, 2017).

Ils sont devenus, en effet, totalement indispensables pour tout travail de recherche et de développement dans l'immense domaine des *linked and open data* (en français : données ouvertes et liées). Lier une donnée à une autre veut dire d'une manière très proche de la vision structurale du langage : expliciter sens d'une donnée (relatif à un contexte d'application, d'usage) en référence à la *position* de la donnée dans une *configuration* d'autres données avec lesquelles elle entretient des relations.

Pour prendre l'exemple d'une application terminologique, le sens d'un terme – sa position particulière par rapport aux autres termes constituant une terminologie – est, entre autres, précisé par son degré de *spécialisation* (sa place dans une hiérarchie *hypéro-/hyponymiques*), par ses rapports de *synonymie partielle* avec d'autres termes, par le ou les *référents* thématisés dans ses usages. Avec comme modèle de référence une telle structure, on peut qualifier un terme concret comme un terme qui est :

1. synonyme (ou partiellement synonyme) à tel(s) autre(s) terme(s) ;
2. l'hyperonyme de tel(s) autre(s) termes ;
3. l'hyponyme de tel(s) autre(s) termes ;
4. utilisé pour exprimer, parler de tel(s) référent(s) ; etc.

C'est une question empirique de savoir si cette structure, telle quelle, est suffisamment riche, explicite… pour pouvoir servir de modèle à la construction d'une base terminologique. En tout cas, on voit bien que la structure, elle, forme un *réseau*, un *graphe* composé d'un *ensemble de triplets*.

Citons ici la théorie des graphes conceptuels développée par John Sowa dans son livre *Conceptual Structures in Mind and Machine* (Sowa 1984). La théorie des graphes conceptuels de J. Sowa propose un cadre remarquablement sophistiqué, formellement explicite et simultanément d'un usage assez intuitif pour structurer et représenter les connaissances d'un domaine dans le cadre d'un projet concret. Les graphes conceptuels sont en effet, largement utilisés dans les recherches en IA, TAL[2] et dans le web sémantique.

---

[2] Nous voudrions citer ici une collaboration scientifique que nous avons pu coordonner au CNRS et ensuite à l'Inalco avec une équipe de chercheurs chez IBM France à la fin des années 80 et au début 90 autour de Jean Fargues, Jean-Pierre Adam et alii sur l'analyse sémiolinguistique et la représentation de corpus textuels (notices pharmaceutiques, presse financière, guides touristiques) et lexicaux à l'aide de graphes conceptuels pour expérimenter la compréhension et la génération semi-automatique de textes spécialisés (Fargues et al. 1986 ; Fargues 1991 ; Stockinger 1989, 1992, 1995).

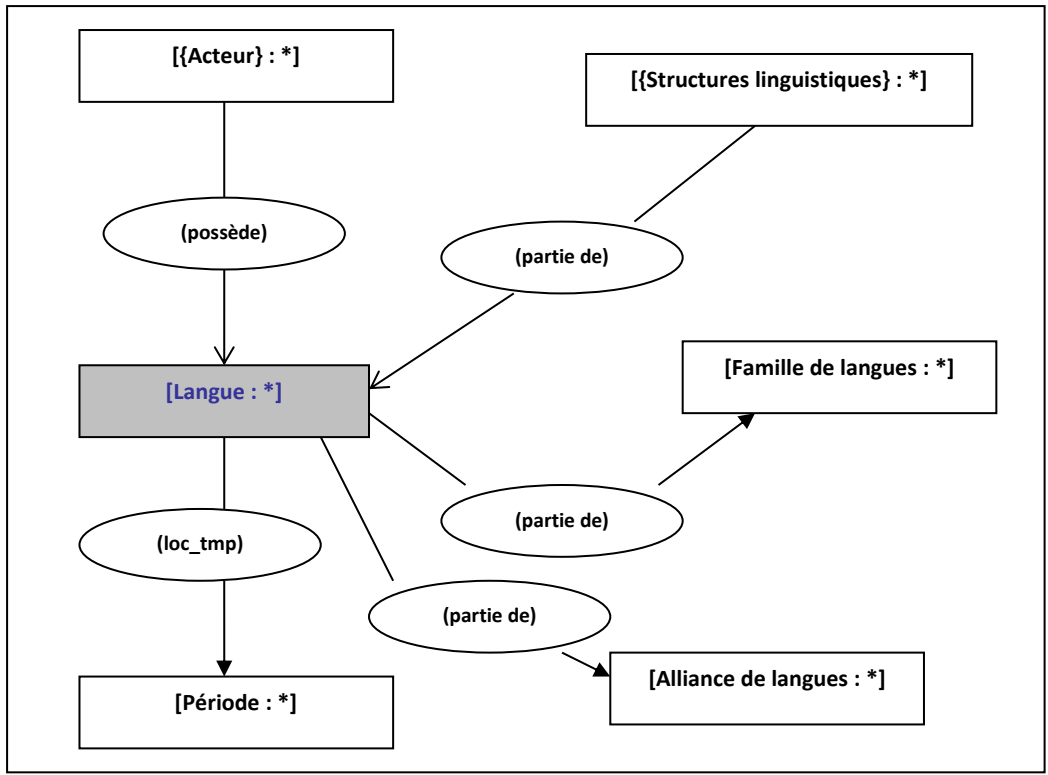

(**Figure 1 :** Graphe représentant une vue sémantique sur l'objet « Langue »)

La figure 1 nous montre un graphe de connaissance qui représente une vue sémantique très simple, un modèle de l'objet « Langue »[3]. Ce modèle fait partie d'un ensemble – d'une « bibliothèque » de modèles – dont l'objectif, la raison d'être est de permettre la description, l'analyse, l'enrichissement de corpus textuels (ici : audiovisuels) documentant une partie de la culture linguistique d'un acteur social (d'une communauté, d'un peuple, etc.). Ce graphe spécifie en gros que la description du contenu d'une vidéo consiste en une description ou en une analyse de plusieurs thématisations dans cette vidéo :

1. la thématisation de l'acteur concerné (de la communauté, du peuple) ;
2. la thématisation de la langue concernée ;
3. la thématisation de l'appartenance de cette langue à une famille ou bien alliance linguistique ;
4. la thématisation de la ou des structures linguistiques (phonologie, grammaire, lexique, etc.) ;
5. la thématisation de la localisation temporelle (d'une époque, d'une période).

---

[3] Ce graphe fait partie d'une « bibliothèque de graphes » se basant sur une ontologie de domaine consacrée à la diversité linguistique et culturelle et qui a été élaborée entre 2006 et 2009 dans le cadre d'un projet ANR (projet SAPHIR ) et d'un projet européen (projet LOGOS ). On trouvera dans notre e-publication « The PCI – Ontology » (Stockinger 2007) une présentation partielle de l'ontologie et de la bibliothèque des graphes. Ajoutons que l'ontologie en question fait partie aujourd'hui d'un projet d'ontologie plus générale dont l'acronyme est ASA (Atelier de Sémiotique Audiovisuelle ; Stockinger 2015) et qui est fournie avec la plateforme sémantique OKAPI de l'INA pour servir à la description de corpus audiovisuels notamment scientifiques (Lalande 2015 ; Beloued 2017).

Des vidéos ne fournissant pas ce genre de thématisations ne pourront pas être décrites par ce modèle. D'autres aspects importants d'une vidéo (par exemple, les stratégies de thématisation, les modalités d'expression des thèmes, etc.) ne pourront pas non plus être décrits par ce même modèle. On voit bien que malgré sa structure simple, le modèle est assez contraignant, ne permet de considérer qu'un type très particulier de contenus audiovisuels et exclut énormément d'éléments sémiotiques qui jouent un rôle dans l'identité d'une vidéo concrète. Mais, tout cela est bien intentionnel et est justifié dans le cadre d'un projet d'analyse d'archives audiovisuelles (Stockinger 2012).

En considérant maintenant les différents éléments du modèle représenté par la figure 1, on voit bien que le graphe en question est composé par des *boîtes* et des *lignes* entre les boîtes :

1. Les *boîtes* correspondent aux sommets ou nœuds dans un graphe et contiennent des *concepts* (= les mots en majuscule entre crochets), des *ensembles de concepts* (= les mots entre accolades à l'intérieur des crochets) et des *individus* ou *des valeurs* (= la petite étoile à droite du double point). Entre concept et valeur, il existe une relation *d'instanciation* ou de *subsomption* représentée par le double point « : ». Un individu, une valeur est une instanciation, un représentant, une occurrence, etc. d'un ou de plusieurs concepts ; le concept qualifie le *type sémantique,* le *sens (partiel)* d'un ensemble d'individus, d'occurrences, etc.

2. Les *lignes* entre les boîtes sont orientées et correspondent aux arêtes dans un graphe. Elles représentent les rapports sémantiques qu'entretiennent les concepts entre eux (= les mots en minuscule). Un rapport sémantique est toujours *orienté* : il y a un concept-*source* et il y a un concept-*cible*.

Sans vouloir entrer ici dans trop de détails, on voit bien que cette conception n'est pas très loin de celle – structurale – de la *configuration (sémantique* ou *conceptuelle)* tout en apportant à cette dernière un appareillage formel explicite et assez contraignant.

## 3) Questions relatives à la modélisation d'un domaine de connaissance

L'élaboration d'un ou, plutôt, d'une *bibliothèque* de modèles sémantiques qualifiant un domaine ou un objet de connaissance à l'aide de graphes conceptuels renvoie à toute une série de problématiques spécifiques dont nous allons rapidement considérer quelques-unes.

Une *première problématique* est constituée par celle de l'identification et de la définition d'un *canon*, d'une *base canonique* de concepts dont on fait l'hypothèse qu'il nous permet de tenir compte d'un domaine ou objet de connaissance donné. Le modèle représenté dans la figure 1 suggère en effet, que les six concepts font partie d'un tel canon dans le cadre d'un projet d'analyse d'un corpus de ressources audiovisuelles thématisant la culture linguistique d'un acteur social. Mais, rien ne dit qu'il s'agit ici d'un *canon complet.* On remarque, par exemple, que ce modèle ne réserve aucune place aux *usages (sociaux, ...)* d'une langue. Ainsi, des thématisations de ce type ne pourront pas être considérées par le modèle en question.

Une *seconde problématique* est concernée par l'identification et l'élaboration d'une *hiérarchie de concepts* s'enracinant dans la base d'un canon conceptuel, pour pouvoir traiter d'une manière plus approfondie un domaine de connaissance. Par exemple, le modèle représenté par la figure 1 suggère que le concept [{Structure linguistique}] est un concept de base qui se déploie en une hiérarchie (ou, plutôt un *treillis*) de concepts comprenant de types plus spécialisés du genre [Lexique], [Grammaire], etc. Intégrer une telle différenciation dans un modèle tel que celui représenté dans la figure 1 signifie que la thématisation de l'objet [{Structure linguistique d'une langue}] dans un corpus de vidéo peut être traitée du point de vue du lexique, de la grammaire, etc. de la langue concernée. Le degré approprié de *généricité* d'un modèle conceptuel dépend en effet, du projet d'analyse et de ses objectifs.

Une *troisième problématique* est réservée à l'identification et à la définition de la *variété* sémantique des *occurrences* ou des *individus* qui instancient un ou plusieurs concepts. Ainsi le concept [Famille de langues] dans la figure 1 n'est pas spécifié davantage en une hiérarchie conceptuelle (permettant d'identifier, par exemple, certains types de regroupements opérés en linguistique comparative) mais *l'ensemble de ses individus* est consigné dans un *thésaurus* de familles, sous-familles, branches et autres regroupements de langues. Concrètement, cela signifie, par exemple, que lorsqu'un analyste souhaite décrire dans une vidéo donnée la thématisation de l'appartenance d'une langue à une famille, il choisira dans le thésaurus la famille ou la branche appropriée.

Une *quatrième problématique* renvoie à la question de l'identification et de la définition des *relations entre concepts* et leur regroupement en une ou plusieurs hiérarchies de relations conceptuelles. Un point important ici est celui de l'explicitation des *contraintes sémantiques* de chaque relation, c'est-à-dire le fait que deux concepts quelconques ne peuvent pas contracter n'importe quelle relation. Ainsi la relation de *localisation temporelle* (= *loc-tmp*) dans la figure 1 ne peut être contractée qu'entre d'une part, un *perdurant* (par exemple, un processus, une activité) ou un *endurant* (par exemple, dans notre cas, un objet culturel, une langue) et d'autre part, une *région temporelle* (dans notre cas, une époque ou une période). Une relation d'appartenance (représentée dans la figure 1 par l'abréviation « partie de ») ne peut être contractée qu'entre deux endurants indépendants, comme entre [Langue] et [Famille de langues] (figure 1). On voit bien qu'il existe une forte interdépendance entre type de concepts et type de relations, ce qui rend le travail du design, du contenu, de la modélisation d'un domaine de connaissance encore plus ardu.

Comme le suggère d'ailleurs la distinction entre *concept* et *instance* ou *valeur d'un concept* ainsi que celle entre rapport entre concepts vs rapports entre individus/valeurs d'un concept, on distingue entre *graphes génériques* (relations entre concepts) et *graphes d'individus* (relations entre valeurs ou occurrences). Les graphes génériques représentent des modèles de connaissance qui qualifient un type de situations, d'événements, d'actions, etc. ; les graphes d'individus représentent, comme son nom l'indique, des connaissances locales relatives à une instance ou une occurrence d'un concept. Pour prendre un exemple simple, le graphe générique de l'objet « Auteur » qualifie ce rôle social dans un réseau de notions telles que « Nom (d'auteur) », « Ouvrages (de l'auteur) », « Lieu d'origine (de l'auteur) », « Date de naissance », « Date de décès », etc. Étant donné l'objectif d'un projet et le contexte d'application ou d'usage, ce modèle peut, bien sûr, comporter tout un ensemble d'autres notions. Le point important à retenir ici, c'est le fait que ce modèle ne qualifie pas un auteur en particulier, mais le rôle social « Auteur » *en tant que tel* (et toujours, bien sûr, en relation à un projet d'analyse) ! En revanche l'assertion que *Victor Hugo* est l'auteur, entre autres, des *Misérables*, qu'il est né en *1802* à *Besançon*, qu'il est décédé en *1885* à *Paris*…, cette assertion est organisée et exprimée sous forme d'un graphe d'individus : les expressions « Victor Hugo », « Besançon », « Paris, « 1802 »,

« 1885 »… sont considérés ici comme des individus (en linguistique de corpus, un type d'individus utilisé abondamment sont, par exemple, les *entités nommées*).

Revenons à notre observation ci-dessus, à savoir que le modèle générique dans la figure 1 n'intègre pas un concept du genre [Usage (social d'une langue)]. Tel quel, il ne pourra pas être utilisé pour analyser, dans un corpus de vidéos, des thématisations documentant le rôle d'une langue dans la vie sociale d'un acteur. Cette limitation nous sensibilise à la double problématique suivante :

1. Celle que nous avons déjà rencontrée ci-dessus (voir *seconde problématique*), de l'identification et de la description de concepts (et, éventuellement, de relations) permettant de considérer une variété plus importante de problèmes de description et d'analyse de corpus audiovisuels : d'autres thématisations (telle que celle justement relative aux usages sociaux d'une langue), de stratégies rhétoriques et visuelles, etc.
2. Celle – nouvelle — de construire non pas de modèles statiques (tel que celui représenté dans la figure 1), mais des *modèles pouvant s'adapter aux besoins* variés de l'analyste d'un corpus audiovisuel.

La première question renvoie au fait que le design du contenu, de la sémantique d'un domaine de connaissance ne se réduit pas à un seul graphe constitué de quelques concepts et relations. Par exemple, le graphe représenté dans la figure 1 forme un réseau sémantique ou conceptuel (une configuration particulière) qui se tisse entre les six concepts qui font partie d'une hiérarchie conceptuelle et un ensemble d'individus réunis en un thésaurus. Le, pour parler ainsi, « tissage », d'une part, entre ces quelques concepts et, d'autre part, avec un nombre indéfini d'individus, est rendu possible grâce aux relations conceptuelles identifiées et définies par le design sémantique du domaine de connaissance. Autrement dit, un modèle conceptuel ou sémantique d'un domaine de connaissance repose sur :

> 1) une *hiérarchie de concepts* ;
> 2) une *hiérarchie de relations* ;
> 3) des *individus* pouvant être réunis sous forme de thésaurus, de terminologie, de lexique, etc. ;
> 4) de *configurations* caractéristiques entre ces trois premières unités, autrement dit : de réseaux sémantiques représentés sous forme de *graphes conceptuels*.

Considérons rapidement la seconde question posée ci-dessus. Elle nous renvoie en effet, à une nouvelle – *cinquième* — *problématique* dans le design du sens d'un domaine de connaissance.

Celle-ci met l'accent sur le fait qu'un réseau conceptuel, qu'un modèle tel que celui représenté dans la figure 1 devrait s'adapter (dans la mesure du possible) aux besoins, aux souhaits de l'analyste et lui permettre de décrire, d'analyser, d'annoter des thématisations relatives à l'usage social d'une langue (pour rester avec notre exemple). En effet, des modèles dépourvus de cette dimension d'adaptation nous mettent dans une situation bien fâcheuse, à savoir de devoir préciser, pour chaque analyste, pour chaque usage particulier, un modèle de connaissance à part entière. Cette perspective n'est guère envisageable. Or, la théorie des graphes conceptuels nous permet justement de considérer (sinon de résoudre efficacement) cette adaptabilité d'un modèle de connaissance à une diversité plus ou moins grande de besoins et d'usage (tout dépend, bien évidemment, de la qualité du design conceptuel du domaine de connaissance). Ainsi, dans son ouvrage *Conceptual Structures* (1984), John Sowa définit un ensemble de règles de *formation* et de *transformation de graphes* ; de procédures *d'expansion* d'un concept en un graphe entier et, réciproquement, de *condensation* d'un graphe en un concept ; de *projection* d'un graphe sur un autre et d'*abstraction* d'un graphe à partir d'une base de graphes ; d'*enchâssement* d'un graphe dans un autre, etc. (voir aussi Fargues et al. 1986 ; Fargues 1991). Toutes ces règles et opérations contribuent à rendre un modèle donné plus dynamique, plus adaptable aux besoins variés d'un analyste ou d'une communauté d'analystes de corpus audiovisuels.

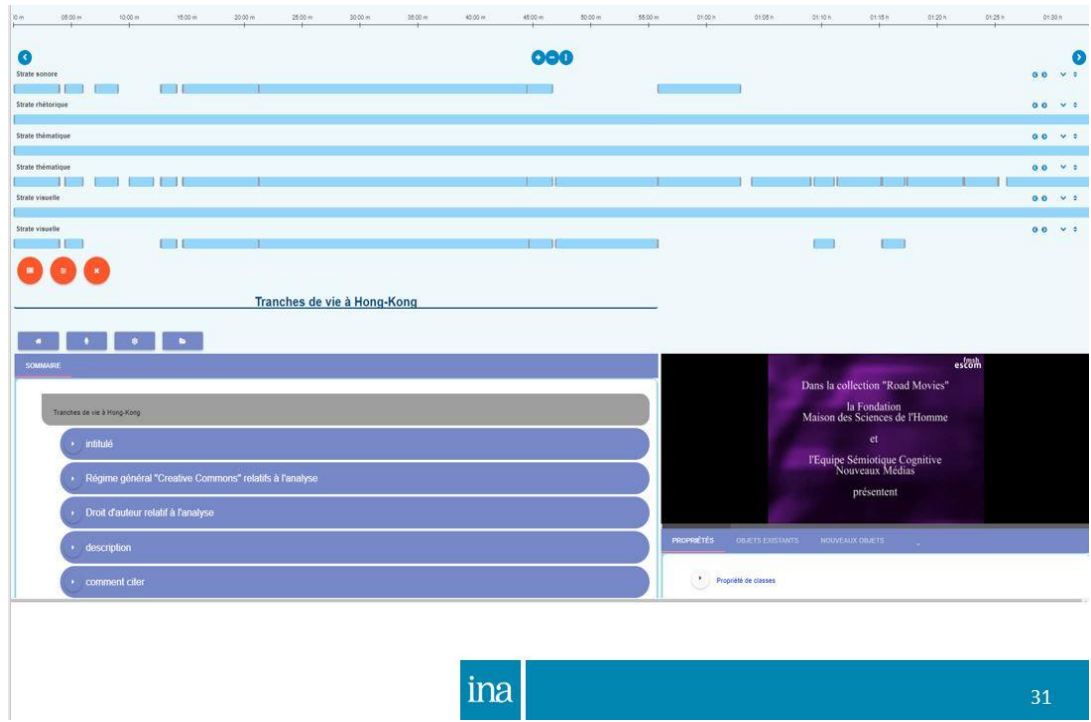

(**Figure 2 :** Formulaire dynamique pour une analyse en ligne d'une ressource audiovisuelle ; environnement de travail OKAPI ; S. Lalande et A. Beloued 2017)

Concrètement parlant : à partir d'un modèle présenté, par exemple, sous forme d'un formulaire dynamique (figure 2), l'analyste doit, en effet, pouvoir, s'il le souhaite, de l'exploiter dans des directions les plus diverses afin de produire des descriptions et des analyses adaptées à ses besoins, à ses objectifs. Néanmoins, ces utilisations variées se réalisent toujours dans le même cadre sémantique, c'est-à-dire en respectant les contraintes spécifiques au modèle sémantique. Autrement dit, l'enjeu central ici est de conjuguer, d'une part, souplesse, *adaptabilité* du modèle d'un domaine de connaissance et, d'autre part, *respect de ses contraintes sémantiques et logiques* – tout en tenant compte, bien entendu des spécificités et des objectifs d'un projet d'analyse de corpus textuels (dans notre cas, audiovisuels).

L'environnement de travail OKAPI (Open Knowledge-based Annotation and Publishing Interface) développé par Steffen Lalande et Abdelkrim Beloued de l'INA (Beloued et al 2017) essaie de répondre à cet enjeu en fournissant des outils et services web pour la modélisation de domaines de connaissance, l'analyse de corpus audiovisuels documentant un domaine de connaissance et la publication/republication de corpus analysés. L'ensemble des services, des outils, ressources sémantiques, etc. fournis par cet environnement est fondé sur la théorie des graphes conceptuels et les technologies et les langages du web sémantique.

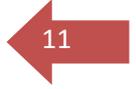

Bien entendu, pour qu'un modèle réponde effectivement aux désirs variés d'un analyste, nous n'avons pas seulement besoin du formalisme des graphes conceptuels ou des langages du web sémantique, mais aussi d'une bonne théorie, d'un bon modèle sémantique du domaine de connaissance. Or, l'élaboration, le design d'un bon modèle d'un domaine de connaissance ne se réduit pas seulement à sa *valeur empirique*, ne se réduit pas seulement, pour reprendre notre exemple du graphe dans la figure 1, à sa capacité de tenir compte d'une certaine diversité de thématisations dans un corpus audiovisuel documentant le patrimoine linguistique d'un acteur social. Ainsi, nous rencontrons ici une *sixième problématique* qui est celle de la *portabilité* et de l'*interopérabilité* du modèle. Portabilité et interopérabilité ne sont pas identiques, mais fortement liées :

*1 - Portabilité* veut dire, succinctement, l'utilisation ou, plutôt, la réutilisation des parties du modèle sémantique d'un domaine de connaissance dans d'autres projets d'analyse. Dans une publication récente (Stockinger 2015), nous avons montré que malgré leur très grande diversité thématique, les enregistrements audiovisuels documentant la recherche en SHS (sciences humaines et sociales) se caractérisent quand même partiellement par des structures très répétitives. Sans rentrer ici dans plus de détails, on peut constater que la narration explicative ou historique d'une recherche dans un domaine se fait à l'aide d'un ensemble de *topoï* qui sont relativement indépendants d'un champ disciplinaire. Autrement dit, des modèles explicitant ce genre de topoï peuvent être réutilisés tels quels ou moyennant quelques modifications dans des projets d'analyse de corpus audiovisuels qui documentent des champs disciplinaires fort variés.

*2 - Interopérabilité* veut dire tenir compte des ressources métalinguistiques (concepts, modèles, thésaurus, terminologies, etc.) déjà existantes et utilisées dans des projets d'ingénierie documentaire ou autres. Par exemple, nous avons parlé ci-dessus d'un thésaurus de familles de langues qui fournit le champ des individus, des valeurs du concept [Famille des langues]. De même, nous avons expliqué que le concept [Structure linguistique] se différencie par une hiérarchie de concepts plus spécialisés et que chaque concept dispose de sa variété sémantique d'individus ou d'occurrences. Le thésaurus des familles, sous-familles, branches et autres regroupements de langues se nourrit ainsi directement de la liste *Language Families* de l'Ethnologue[4] qui constitue une référence en la matière et auquel se réfère bon nombre de projets autour du monde. A son tour, la différenciation du concept de base [Structure linguistique] et l'assignation à chaque concept d'un champ de valeurs approprié se fait en référence à l'ontologie GOLD (General Ontology for Linguistic Description)[5] qui est une des principales références en la matière. D'autres références que le modèle dans la figure 1 considère, sont *RAMEAU* (le Répertoire d'autorité-matière encyclopédique et alphabétique unifiée)[6] de la Bibliothèque nationale de France, le thésaurus de l'UNESCO[7] ou encore le Thesaurus for Social Sciences (*TheSoz*)[8]. La mise en relation des concepts du modèle dans la figure 1 avec ceux qui font partie des références citées rend (partiellement) interopérable le modèle avec les thésaurus, terminologies et autres ontologies déjà existants[9].

Un des intérêts de cette interopérabilité est simultanément de garantir un accès plus généralisé, plus probable également à une ressource textuelle (dans notre cas : audiovisuelle) et de *restituer le champ intertextuel* pour une ressource donnée qui dépasse de loin le périmètre d'un corpus de données analysées et les annotations d'un analyste. Autrement dit, si nous analysons avec le modèle de la figure 1 un entretien avec un linguiste ayant travaillé sur une langue amérindienne telle que le guarani, il nous collecte à travers les différents moteurs utilisant les mêmes références que les siennes (DBpedia, RAMEAU, GOLD, TheSoz, etc.) *tout un corpus de liens vers de données potentiellement intéressantes* pour mieux comprendre le contenu thématisé dans notre vidéo. Dans un travail de sélection et de structuration éditoriale, on peut ainsi en effet, offrir à des communautés de lecteurs des *graphes documentaires* ou *textuels* reliant un certain nombre de données textuelles avec la vidéo principale.

---

[4] https://www.ethnologue.com/browse/families
[5] http://linguistics-ontology.org/
[6] http://data.bnf.fr/
[7] http://vocabularies.unesco.org/browser/thesaurus/fr/
[8] http://lod.gesis.org/thesoz/de.html
[9] Pour plus d'informations et d'explications, cf. le portail du projet communautaire DBpedia (https://wiki.dbpedia.org/) visant à expliciter, à structurer les informations web sous forme d'un *open knowledge graph* …

Les graphes réunissant un tel champ intertextuel autour d'une ressource vidéo principale peuvent être *figés* (le champ intertextuel n'évolue pas) ou *ouverts* (le champ intertextuel évolue suivant les productions intellectuelles pouvant intéresser la vidéo principale). Les graphes peuvent être de nouveau *non adaptables* (les contours, le périmètre, l'organisation du champ intertextuel ne varient pas) ou *adaptables*, *dynamiques* (le champ intertextuel proposé s'adapte aux besoins, aux désirs, compétences, etc. de l'utilisateur (lecteur). En ajoutant encore le fait que les graphes eux-mêmes peuvent être l'objet de toutes sortes de *stratégies de visualisation* (qui ne sont pas obligatoirement cantonnées dans celle se limitant de rendre un graphe à l'aide de petites boîtes et de lignes entre boîtes …), on peut avoir une idée de la puissance cognitive potentielle des recherches et des développements actuels en ré-ingénierie de corpus de données.

## 4) Distinction entre théorie formelle de représentation et théorie empirique

Terminons notre rapide discussion du design au sens d'un processus d'explicitation de visions d'un domaine de connaissance à l'aide de modèles sémantiques représentés sous forme de graphes conceptuels.

Retenons que la théorie des graphes conceptuels définit, en effet, un *langage de représentation de connaissance* qui possède son *« lexique »,* sa *« grammaire »* et ses *signes d'expression*. Ce langage est formellement défini. John Sowa montre comment exprimer un graphe dans la logique des prédicats et des logiques modales avec la logique temporelle. Un graphe conceptuel n'est pas une simple représentation graphique, un simple aide-mémoire visuel pour organiser des idées émises, par exemple, dans une réunion, etc. En revanche, le langage des graphes conceptuels peut transformer un simple schéma graphique utilisé comme aide-mémoire visuel en un véritable outil de structuration de connaissance !

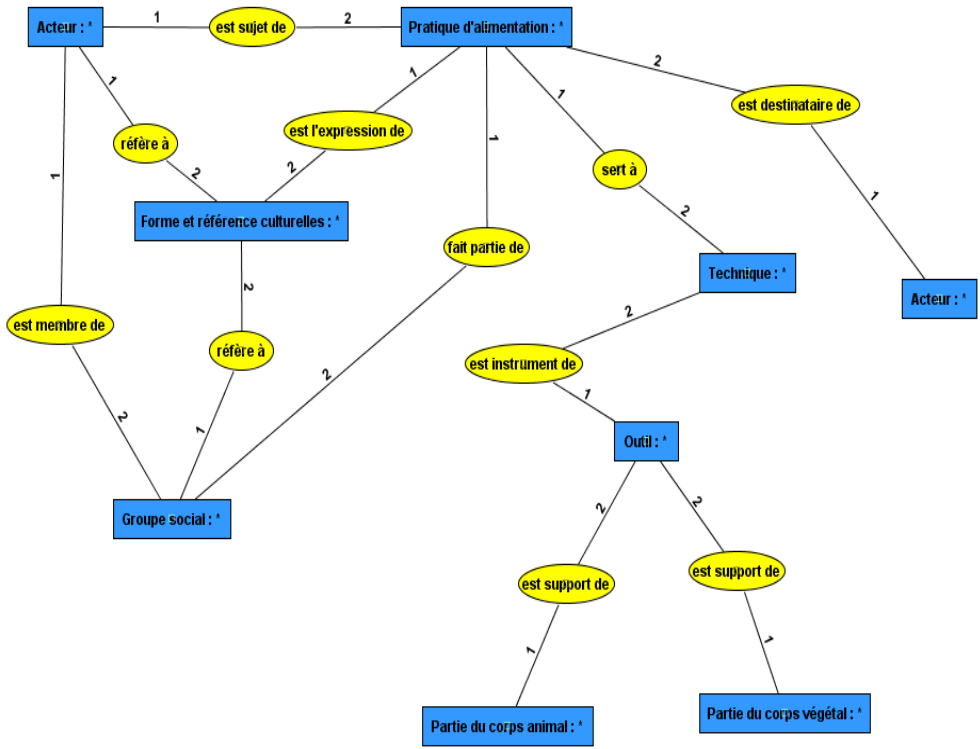

(**Figure 3 :** Graphe représentant une vue sémantique sur l'objet « Pratique d'alimentation »)

Le point crucial ici est cependant le suivant : pour utiliser d'une manière intéressante les graphes conceptuels, il faut déjà avoir une *théorie du domaine* qu'on veut représenter, c'est-à-dire pour revenir à notre propos développe dans le chapitre précédent, il nous faut un modèle ou une vision du domaine. La théorie des graphes aide « seulement » à structurer et à organiser cette théorie, cette *structure des connaissances*. Il faut donc distinguer entre deux théories :

1. une théorie formelle de représentation (c'est-à-dire les graphes conceptuels) ;
2. une théorie du domaine (une théorie empirique).

L'objet qualifié et représenté par un réseau ou un graphe varie, comme déjà dit, d'un projet à un autre. Ainsi, dans le cadre d'un projet d'une base de données terminologiques, l'objet visé est la structure linguistique de l'ensemble des termes composant la terminologie, la signification référentielle des termes ainsi que le ou les contextes d'usage de la base. Dans le cadre d'un projet de compréhension et de génération d'un corpus de notices pharmaceutiques, l'objet visé est d'une part, la structure de ce genre de textes et le domaine qu'ils documentent ainsi que d'autre part, de nouveau les objectifs (le cadre d'exploitation, d'usage) de ce système de compréhension/génération. Dans le cadre d'un projet de valorisation d'une archive audiovisuelle de recherche, l'objet visé est d'une part l'univers sémiotique (les genres, les thèmes, le plan de l'expression…) du fonds audiovisuel et d'autre part, lesusages possibles d'une archive.

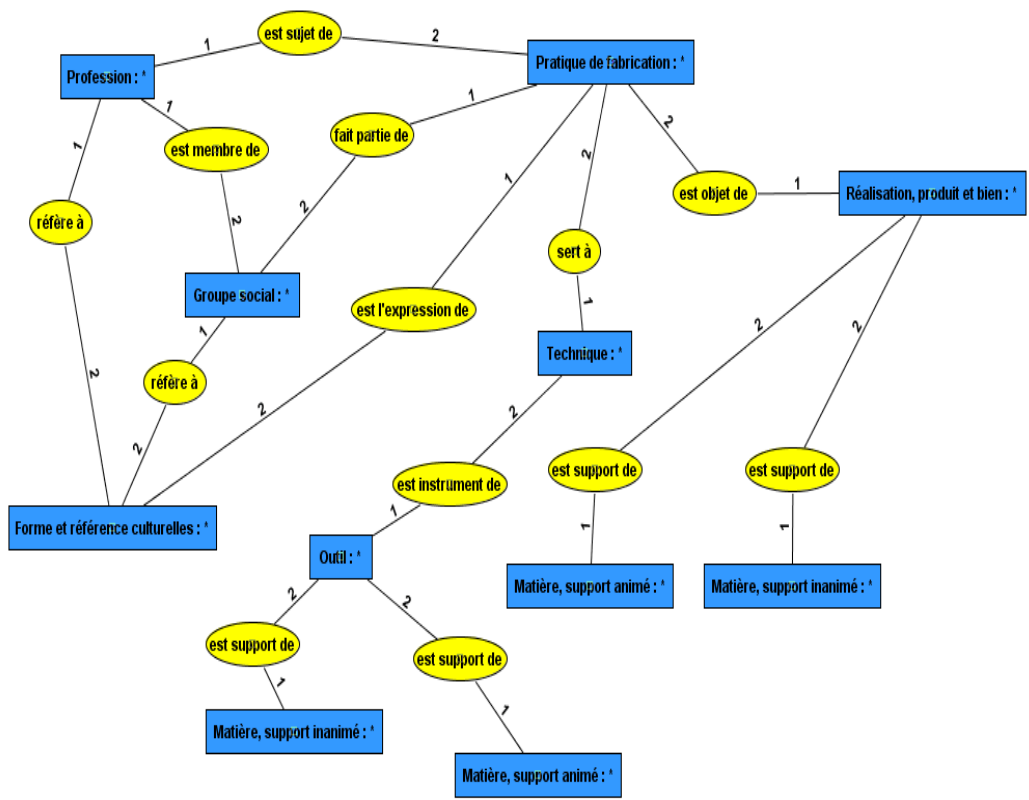

(**Figure 4 :** Graphe représentant une vue sémantique sur l'objet « Pratique de fabrication »)

Considérons encore rapidement les deux figures 3 et 4. Il s'agit de deux captures de graphe de connaissance élaborées dans le cadre d'un projet ANR intitulé SAPHIR (2006-2009) qui portait sur la republication (le ré-ingénierie) sémantique de corpus audiovisuels. Les deux captures représentent des exemples d'une modélisation du domaine de connaissance « Patrimoine culturel (matériel et immatériel) indigène » (c'est-à-dire des peuples autochtones ou indigènes). La figure 2 représente un modèle conceptuel de l'objet « Pratiques d'acquisition et de préparation de la nourriture » et la figure 3 de l'objet « Pratiques de fabrication et de production d'outils et d'instruments ».

Ces graphes servent, par exemple, à l'analyse de corpus audiovisuels (d'images statiques, de vidéos, d'animations, de cartes, etc.) qui thématisent les deux objets ou domaines en question. Le résultat concret d'une telle analyse consiste, par exemple, en un corpus de segments vidéo et de régions visuelles thématiquement annotés. Ils peuvent être regroupés en différentes classes ou catégories sémantiquement homogènes selon le modèle utilisé, mis en relation pour former de constellations d'éléments documentaires, réédités et republiés selon des formats choisis, etc.

Toujours est-il : les différents concepts (dans les boîtes bleues) sont des « objets d'analyse » et les différentes relations (dans les ellipses jaunes) sont des marqueurs de positionnement entre les différents objets d'analyse. Ensemble, ils forment le *modèle de l'objet* ou *du domaine de connaissance* (dans notre cas : des pratiques d'alimentation et des pratiques de fabrication d'outils). Ces deux domaines font partie, comme déjà dit, d'un domaine plus vaste qui est celui du patrimoine (matériel ou immatériel) des peuples autochtones. Les objets d'analyse et les relations entre les objets sont regroupés en des hiérarchies (ou, plutôt, des treillis) et forment ce qu'on appelle l'ontologie d'un domaine. Un graphe de connaissance, lui, tel qu'il a été représenté par les deux figures 3 et 4, est le résultat d'un processus de sélection d'un ensemble de relations projetées sur les objets d'analyse (autrement dit : les relations sélectionnées tissent des rapports entre les objets d'analyse et les regroupent en une configuration sémantique).

L'objet visé – l'*objet intentionnel* – est donc une *structure de connaissance* – un *modèle* au sens cognitif du terme – que le *modélisateur* (le cogniticien, l'ingénieur de connaissance…) extrait d'un corpus, c'est-à-dire d'un *ensemble de données* (terminologiques, textuelles ou autres) *pertinentes* qui documentent le *domaine choisi de connaissance*. La structure de connaissance – le modèle cognitif – extrait, défini et spécifié est alors représentée sous forme d'un *réseau sémantique* ou *conceptuel*.

## 5) Théorie empirique « ad hoc » et théorie empirique générale

Pour revenir à la notion de *théorie empirique*, il convient de distinguer ici entre des *théories (empirique) « ad hoc »*, et une *théorie (empirique) générale*, cette dernière servant de référence à tout un ensemble de projets de modélisation de domaines de connaissance, d'analyse et de publication de corpus.

Théorie empirique *ad hoc* veut dire, dans notre cas, la construction d'un modèle sémantique à partir d'un corpus de données, par exemple, à partir d'un corpus documentant la culture linguistique d'un acteur (voir le modèle de la figure 1) ou à partir d'un corpus documentant le patrimoine matériel et immatériel de peuples autochtones, de peuples indigènes (voir figures 2 et 3).

Actuellement, nous travaillons avec des collègues des laboratoires DeVisu de l'université de Valenciennes (Sylvie Leleu-Merviel) et GERiiCO de l'université de Lille 3 (Stéphane Chaudiron), dans le cadre d'un autre projet ANR, sur le monde de vie des mineurs dans le Nord de la France[10]. Le modèle conceptuel de ce domaine de connaissance que nous essayons de mettre en place se base, est fondé empiriquement sur un corpus de quelque 300 extraits vidéo qui appartiennent à l'INA et diffusés sur le portail « Mineurs du Monde »[11]. Les principales étapes de ce travail sont :

1. L'analyse comparative du contenu du corpus en question.

2. L'identification et l'explicitation des principales situations pro-filmiques documentées, traitées dans ce corpus de vidéo.

3. Sur la base des situations pro-filmiques identifiées, spécification d'une famille de sujets (c'est-à-dire de *topoï* mis en discours) qui nous serviront – d'une manière semblable aux modèles représentés par les figures 1 à 3 – de modèle, de guide pour localiser, segmenter, classer, analyser, enrichir, etc. des vidéos ne provenant pas obligatoirement du fonds audiovisuel de l'INA mais au contraire, de *tout* fonds audiovisuel (institutionnel, privé, scientifique, etc.). Par ailleurs, ladite famille de sujets peut également servir à produire, à réaliser des vidéos « sur mesure » répondant aux points de vue exprimés par ces sujets.

---

[10] http://www.agence-nationale-recherche.fr/Projet-ANR-16-CE38-0001
[11] http://fresques.ina.fr/memoires-de-mines/

4. La définition de l'ontologie du domaine (les objets, les relations, le thésaurus, les individus…), c'est-à-dire la spécification d'un canon de concepts, des hiérarchies conceptuelles, des individus ou des occurrences, des relations entre concepts ainsi qu'entre un concept et ses occurrences ou valeurs. Ce travail doit tenir compte de certaines exigences formelles propres à la théorie des graphes conceptuels et aux langages ontologiques du web sémantique, de l'existence éventuelle d'une ontologie générique (propre, par exemple, à une théorie empirique *générale* ; voir ci-après) et de l'interopérabilité avec des ressources (métadonnées) qui font partie d'autres ontologies, thésaurus, terminologies, etc.

5. La spécification des modèles d'analyse (famille de sujets ; modèle textuel et de publication).

6. Le développement, le test et la validation des modèles d'analyse et de publication.

7. L'ouverture de l'interface de consultation et du travail éditorial à la communauté d'usagers intéressés.

8. L'amélioration, l'enrichissement de l'ontologie, des modèles d'analyse et de publication.

En examinant d'une manière plus précise le contenu de quelque 80 extraits vidéo, nous avons pu constater l'existence de toute une série de *situations pro-filmiques récurrentes* thématisées dans ces vidéos, dont ces vidéos nous racontent des histoires, nous apportent des témoignages, des explications, etc. Parmi ces situations pro-filmiques récurrentes, nous rencontrons, entre autres, les situations suivantes :

> 1. La mine aussi bien au sens d'un gisement de matières premières (du charbon, dans notre cas), d'une organisation sociale (d'une compagnie, d'une exploitation au sens institutionnel du terme), d'un lieu d'activité, d'une installation technique, etc.
> 2. Le monde de travail du mineur, les activités, l'organisation journalière du travail, les conditions de travail…
> 3. Le cadre de vie (hors travail) du mineur : sa famille, son habitat, ses activités de loisir, etc.
> 4. L'histoire simultanément sociale (conscience de classe, syndicalisme, patronat …), politique (gouvernements successifs, guerres, …), industrielle et *technique.*
> 5. Le cycle des *activités industrielles* typiques (découverte du charbon, exploitation, production, transformation…).

En prenant comme point de départ ces différents types de situations pro-filmiques, le travail de modélisation consiste à identifier les sujets (c'est-à-dire topoï « mis en discours ») qui nous serviront à l'analyse d'une certaine *diversité de thématisations* dans des corpus audiovisuels (c'est-à-dire autres que celui de l'INA !) documentant le monde de la mine et du mineur. Dans le cadre du projet MemoMines, nous avons ainsi identifié, parmi d'autres, les six sujets suivants :

1. Sujet « La mine dans le Nord de la France » (= site, gisement, organisation sociale, histoire).
2. Sujet « Découvrir la mine » (construction, lieu d'activités, installation, techniques et technologies) ».
3. Sujet « Le travail dans la mine » (monde de travail, métier, activité, équipement) ».
4. Sujet « Le monde de vie du mineur » (famille, logement, vie quotidienne, loisir, fêtes …) ».
5. Sujet « Risques au travail, maladies professionnelles » (maladies, accidents…).
6. Sujet « Événements marquants dans l'histoire des mines » (événements politiques, événements sociaux, catastrophes).

Chacun de ces sujets forme un *réseau sémantique générique*, i.e. un *graphe* (souvent fort complexe) entre *concepts* ou *classes* (dans la terminologie rdf/owl). En identifiant les concepts, les individus/occurrences et les relations entre concepts composant le *graphe générique* d'un sujet, nous explicitons sa *structure de connaissance*, c'est-à-dire sa vue, son point de vue sur un certain type de situations qui fait partie de notre domaine de connaissance. Par exemple, la figure 4 nous montre – sous une forme tabulaire – une partie du graphe qui spécifie un type de situations thématisé par le sujet « La mine dans le Nord de la France ». Le sujet « La mine dans le nord de la France » est un *pattern commun* à toutes les vidéos, images… *qui proposent des informations sur une mine (houillère) particulière, appartenant à une compagnie minière et à une certaine époque donnée.* La figure 5 nous montre en quoi consiste ce pattern commun. Il se distingue par un concept-source (ou « tête ») qui désigne l'[Objet « Mine » (au sens d'un lieu)]. Cet objet est défini par le fait qu'il entretient des rapports avec toute une série d'autres objets désignés par des termes tels que [Nom du site (d'une mine)], [Objet « Gisement (Type de-) »], etc. Les rapports sont désignés par des expressions telles « Identifier le nom de … », « Préciser le gisement », etc.

| Concept « tête » | Relation | Concept : *valeur* | Notes |
|---|---|---|---|
| Objet "Mine (lieu)" | Identifier le nom | *Nom (du site)* | Liste Memomines des noms des sites dans le Nord de la France (ex. : *Fosse n°1 des mines de Courrières…*) |
| | Préciser l'époque | Période temporelle à spécifier | |
| | Préciser gisement | Objet "Gisement (Type de -)" | Liste Memomines des gisements (*charbon…*) |
| | Préciser la construction | Objet "Mine (construction industrielle)" | Liste Memomines des constructions (*houillère, mine de fer…*) |
| | Préciser la compagnie exploitante | Exploitation minière | Liste Memomines des compagnies minières (*Compagnie des mines de Courrières…*) |
| | | | |

(**Figure 5 :** Extrait d'un graphe « tabulaire » représentant une partie du sujet « La mine dans le Nord de la France »)

Les différents concepts et relations font partie de l'ontologie du domaine, soit en tant qu'entités participant à la base canonique de l'ontologie, soit comme membre d'une hiérarchie conceptuelle. L'ontologie du domaine est complétée par les individus ou occurrences regroupés en un thésaurus du domaine. La figure 6 nous montre un petit extrait de la base canonique de l'ontologie du domaine MemoMines.

|   | Classe (= Objet d'analyse du domaine MemoMines) | Définitions/Remarques | Superclasse ASA |
|---|---|---|---|
| 1 | Objet « Gisement » | Concentration de ressources naturelles (charbon…) | Objet "Formation non-vivante (Type de -)" |
| 2 | Objet « Mine (lieu) »<br><br>Label alternatif : Objet « Carrière minière » | La mine au sens d'un lieu d'activité industrielle | Objet "Lieu d'activité industrielle (Type de-)" |
| 3 | Objet « Mine (construction industrielle) » | La mine considérée comme un ensemble de bâtiments et d'installations<br>Objet réunissant les différents types de complexes industriels (houillères…) et les éléments qui le composent (châssis à molettes…) | Objet "Groupement de constructions (Type de-)" |
| 4 | Objet "Métier de la mine" | Les divers métiers (haveur, herscheur…) réunis dans un petit thésaurus | Objet "Métier de l'industrie (Type de -)" |
| 5 | Objet "Instrument du domaine minier" | Les divers instruments et outils du mineur/en usage dans une mine (berline, balle, astiquette…) réunis dans un petit thésaurus. | Objet "Instrument par domaine d'activités (Type de -)" |
| 6 | …. | | |

(**Figure 6** : Extrait de l'ontologie du domaine MemoMines)

Cet exemple nous montre que la spécification, la définition d'une structure des connaissances, c'est-à-dire d'un modèle de connaissance d'un domaine, est un travail assez complexe exigeant une grande rigueur, pas seulement théorique et méthodologique, mais aussi et avant tout *empirique*.

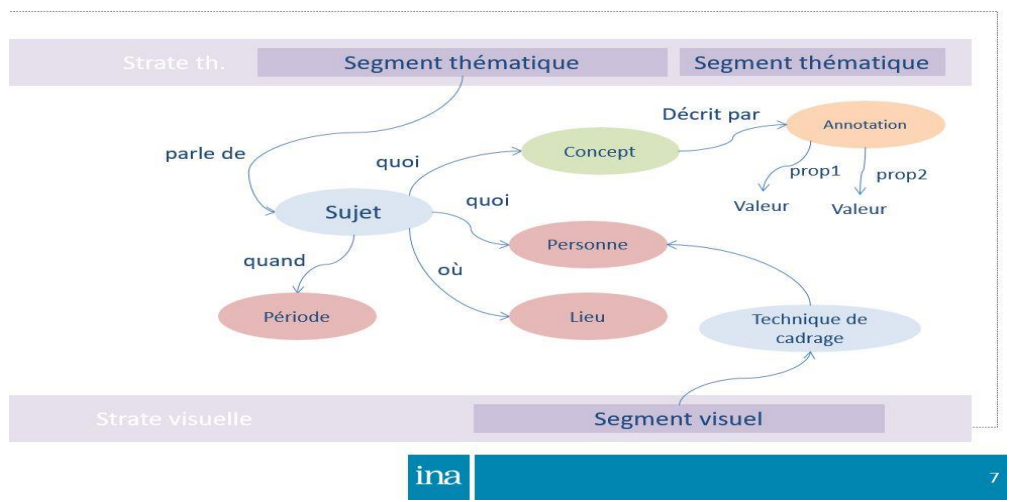

(**Figure 7** : Extrait du modèle général de la donnée textuelle servant de référence à l'élaboration du domaine ASA (Atelier de Sémiotique Audiovisuelle) ; représentation conçue et réalisée par Steffen Lalande, INA, 2017)

L'ontologie du domaine Memomines et ses différents sujets sont un exemple d'une *théorie empirique ad hoc* qui, dans notre cas, prend place dans une *théorie empirique générale* fournie par l'ontologie ASA (Atelier de Sémiotique Audiovisuelle). L'ontologie ASA a été progressivement développée et expérimentée depuis une dizaine d'années, entre autres grâce à trois projets de recherche ANR[12]. *Théorie empirique générale* veut dire, ici, réutiliser partiellement dans le cadre de Memomines des modèles déjà existants ainsi que, réciproquement, intégrer Memomines dans un *cadre conceptuel commun plus large*, réutiliser des modèles Memomines et d'autres entités dans de nouveaux projets, etc.

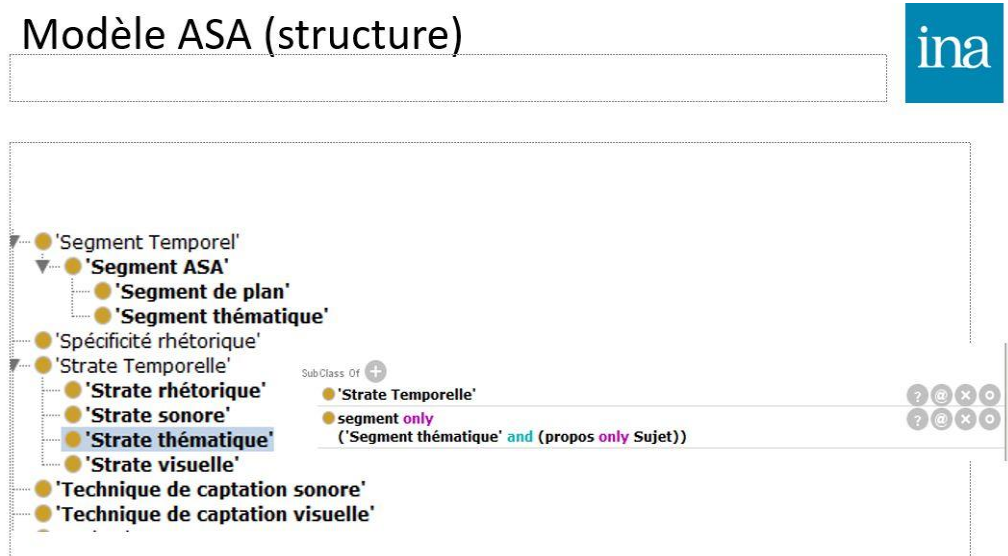

(**Figure 8** : Modèle cœur de connaissance ASA sur Protégé ; réalisé par Steffen Lalande, INA, 2017)

Répondre à ce genre d'exigences est l'objectif de l'ontologie ASA (Atelier de Sémiotique Audiovisuelle), déployé sur l'environnement OKAPI – un environnement conçu et développé par Steffen Lalande et Abdelkrim Beloued de l'INA (Lalande et al. 2015 ; Beloued et al 2017). Cette ontologie est testée sur des corpus audiovisuels thématiquement très variés provenant du fonds du programme ESCoM-AAR (2001 – 2016)[13] ainsi que progressivement généralisée et enrichie pour pouvoir considérer des corpus audiovisuels autres que scientifiques. La figure 7 nous montre un extrait du modèle de base (du « cœur ») qui sous-tend toute cette entreprise ASA. Ce modèle considère un texte lato sensu (ici : une vidéo) comme étant constitué, entre autres, par un ensemble de *strates* (thématiques, rhétoriques, visuelles, etc.). Chaque strate est composée d'un ensemble ouvert de segments. L'identification et la description d'un segment se font selon les contraintes propres à une strate : contraintes thématiques pour ceux qui concernent les segments faisant partie de la strate thématique ; contraintes visuelles concernant les segments qui font partie de la strate visuelle, etc.

---

[12] Il s'agit des projets ANR suivants : SAPHIR (2006 – 2009) ; ASA-SHS (2009 – 2012) et Campus AAR (2014 – 2017)

[13] https://hal.campus-aar.fr/

Parmi les contraintes thématiques, on compte, entre autres, les déjà cités, *sujets* ou *topoi* développés, mis en discours dans une vidéo ou un corpus de vidéo. Parmi les contraintes visuelles, on compte, par exemple, les différentes techniques de prises de vue. Entre deux strates d'une vidéo (dans notre cas : entre sa strate thématique et sa strate visuelle) se tissent des interdépendances qui se manifestent, par exemple, par l'usage des différentes techniques de visualisation d'un sujet : angle de vue de la caméra, plans visuels utilisés, mouvements caméra, éclairage, etc. (voir figure 7).

La figure 7 représente le modèle général de l'objet de connaissance « texte » ou « ressource textuelle » lato sensu (y compris donc le texte audiovisuel) sous forme d'une représentation graphique. La figure 8 explicite cette représentation graphique d'une théorie du texte audiovisuel sous forme d'une petite base canonique de concepts et de hiérarchies conceptuelles (le logiciel utilisé pour visualiser cet extrait ontologique d'un modèle du texte/de la ressource textuelle est *Protégé* de l'université de Stanford).

La figure 9 montre, enfin, des extraits d'un ensemble de composants de l'ontologie générique ASA utiliser pour spécifier des modèles d'analyse de vidéos provenant documentant de domaines de connaissance variés : domaines disciplinaires en SHS, patrimoine industriel, patrimoine territorial et communal, etc.

Pour terminer, l'ontologie est un « gros graphe » qui représente le modèle sémantique (ou conceptuel) d'un domaine de connaissance générique ou spécifique. Les arêtes entre concepts ou occurrences, valeurs, individus, etc. proviennent du treillis des relations, les concepts du graphe proviennent du treillis de concepts du domaine et les valeurs ou occurrences proviennent du thésaurus du domaine.

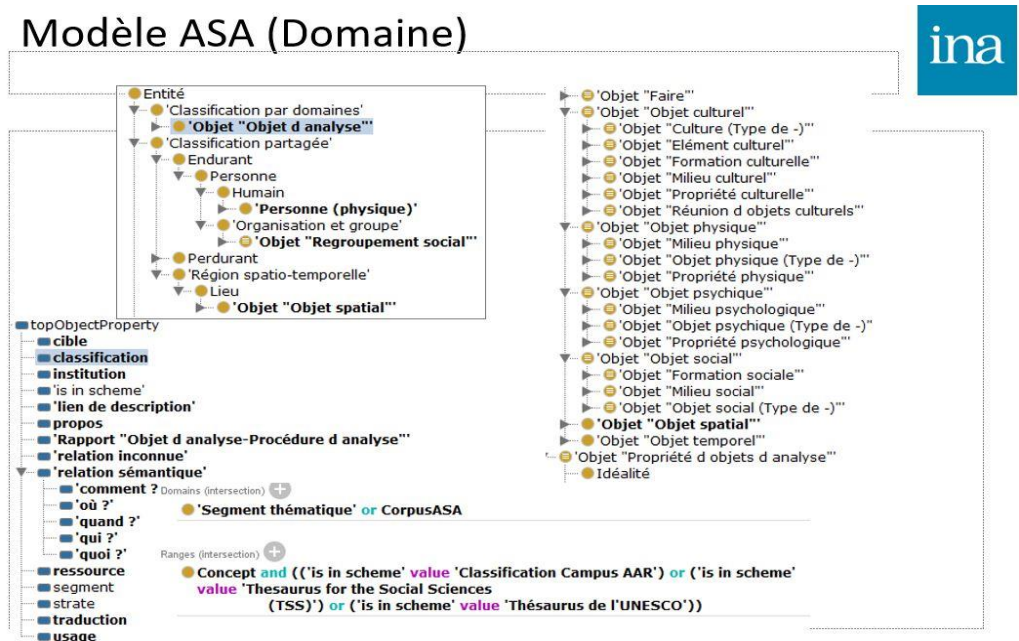

(**Figure 9** : Éléments de l'ontologie générique ASA sur Protégé ;
réalisé par Steffen Lalande, INA, 2017)

Grâce aux propriétés formelles du langage des graphes conceptuels, on peut procéder à tout un ensemble d'opérations de restriction sur les propriétés du graphe afin d'expliciter autant que faire se peut la structure de connaissance que nous souhaitons utiliser pour manipuler des corpus de vidéos : pour les découper, les indexer, les analyser, pour y accéder en tant qu'utilisateur, pour les publier ou republier sous forme de portails, de dossiers à thème, etc. Pour pouvoir réaliser tout cela, il faut également un environnement de travail, un ensemble d'outils, de logiciels, d'applications, etc. fondé sur le web sémantique et le formalisme RDF et OWL (Web Ontology Language). Notre environnement de travail est, comme déjà dit à plusieurs reprises, OKAPI (Open Knowledge Annotation and Publishing Interface), actuellement en cours de développement par Steffen Lalande et Abdelkrim Beloued de l'INA[14].

Tout en nous contentant de ces quelques explications et ces exemples, on voit bien ce que c'est le *design du sens (de la valeur) d'un domaine* :

1. Le design du sens (de la valeur) d'un domaine, c'est la définition d'un modèle de connaissance d'un domaine donné, c'est-à-dire d'une structure de connaissance qui nous est nécessaire pour réaliser les objectifs d'un projet donné.

2. Ce modèle peut être exprimé dans un langage de graphes conceptuels (Sowa).

3. Mais le modèle lui-même, sa structure, est le résultat d'un travail empirique reposant sur la prise en compte d'un corpus de données qui documente le domaine à modéliser ainsi que les objectifs et les contextes d'usage ou d'exploitation.

4. Ce travail empirique présuppose la référence à un cadre théorique qui propose une vision générale et apriorique (mais toujours révisable, bien sûr !) des données à modéliser.

5. Si la structure de connaissance spécifiée doit être exprimée dans un langage de graphes, considérons les contraintes de ce langage (plus ce langage est formellement ou syntaxiquement explicite, plus la structure de connaissance sera explicite, mais plus les contraintes à respecter seront importantes).

6. La qualité du modèle, de la structure des connaissances dépend directement de sa valeur empirique et de son caractère formellement explicite, mais elle dépend également de son interopérabilité et de sa dérivation explicite d'une ontologie fondationnelle.

## 6) Le design au sens de la mise en scène multimodale des métadonnées et des données textuelles

Venons-en maintenant au design compris comme une activité de l'expression, de la mise en scène (visuelle, graphique…) d'une information. Il convient de regarder de plus près quelques points de compréhension plus généraux.

---

[14] Adresse pour visualiser un premier prototype peut être visualisé :
http://gallium.lab.parisdescartes.fr:31480/files/html/analysis.html?activeLeftMenu=menu_liste_analysis

Il y a d'abord *l'expression, la mise en scène « naturelle »* d'un sens, d'un contenu. Typiquement, ce sont les mots d'une langue et, d'une manière générale, tous les signes perceptibles qui forment le milieu signifiant - le textscape (Stockinger 2016) — danslequel l'homme est né et dans lequel il grandit. En fait, ils'agit ici du *monde naturel* dont parle A.J. Greimas (1966, 1976). Le monde naturel, c'est un design *qui ne se donne pas* comme design (intentionnel) d'un sens, mais comme quelque chose qui est là, comme la réalité sensible, primaire, vécue – comme une *nature*. Entendu dans ce sens, le design de l'information se confond avec l'histoire du langage.

Ensuite, nous avons le design compris comme une pratique, comme une technique spécialisée qui sert à un objectif de communication défini en amont à tout *projet de design*. C'est le cas, par exemple, de l'« information design » au sens anglais du terme qui a une histoire plus longue que le très récent « data design ». Comme déjà dit au début de ce travail, l'*information design* est un terme utilisé pour désigner un ensemble de *techniques de visualisation* utilisées depuis au moins les années 1960 en communication scientifique et technique et un peu plus tard aussi, en didactique des sciences. Ces techniques servaient – et servent d'ailleurs toujours – à produire des représentations schématiques, facilement assimilables, de contenus souvent très spécialisés qu'on trouve typiquement dans les manuels techniques et scientifiques. En didactique des sciences et en psychologie cognitive, on utilise des techniques de visualisation analogiques ou métaphoriques pour faciliter l'accès à des connaissances scientifiques ou techniques complexes.

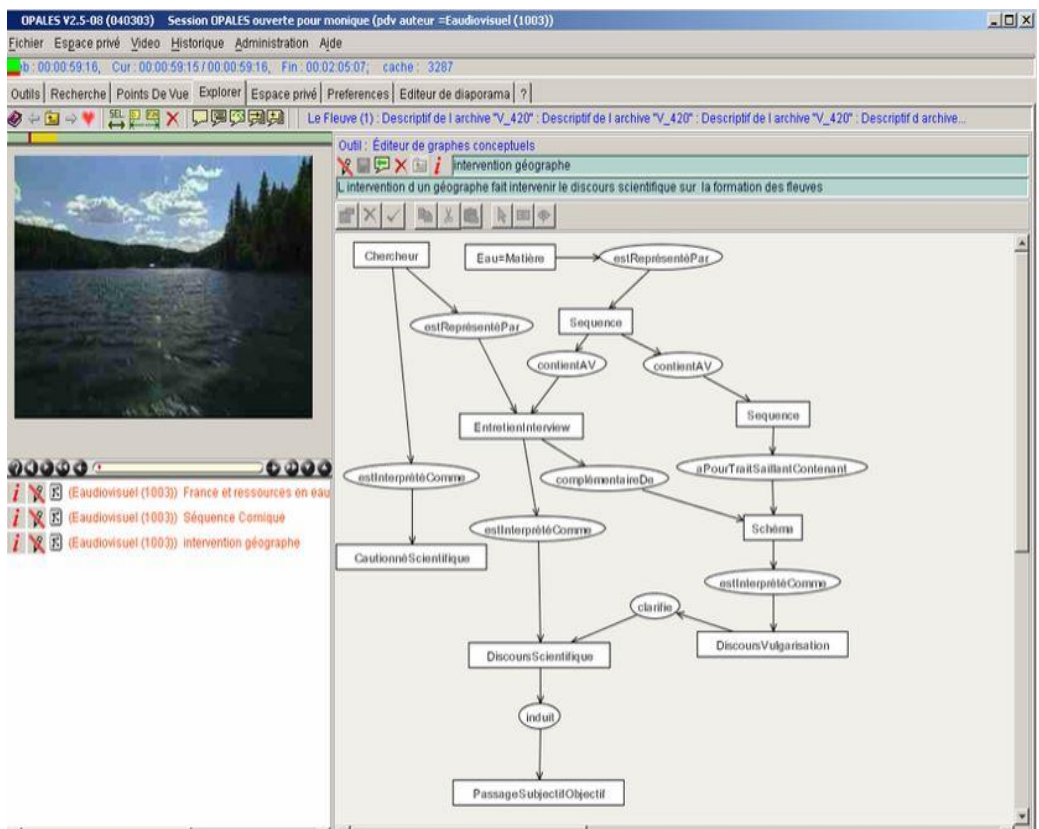

(**Figure 10** : Interface OPALES (2001 – 2003), réalisé par l'INA)

Lorsqu'on parle de la technique de *data visualisation,* de la visualisation de données, on peut également se référer à des choses différentes. Le sens le plus usuel aujourd'hui est celui de la visualisation de grandes masses de *valeurs à variables quantitatives* pouvant être traitées par des méthodes statistiques, probabilistes, etc. Cette visualisation se fait, par exemple, sous forme de graphes se déployant dans un espace 2D ou 3D, de diagrammes à bâtons, de diagrammes circulaires, de cartes statistiques, de nuages de points, etc. Mais, elle peut aussi revêtir des formes plus métaphoriques, imitant un univers eidétique particulier (par exemple, celui des émoticons ou des personnages de fiction)

Un troisième sens est celui de la *visualisation des vues*, des *points de vue* sur des corpus de données. Ces vues, ces points de vue sont donc, en d'autres mots, des *interprétations contrôlées* (contrôlées, plus exactement, par un ou plusieurs *modèles conceptuels*) d'un ensemble de données. Les données, elles, sont typiquement (mais pas exclusivement) des données *textuelles* lato sensu qui documentent un domaine, un objet de référence (tel que celui du monde des mines ou du patrimoine matériel et immatériel d'un acteur social, etc.). En d'autres termes, il s'agit ici de la visualisation *non pas des données*, mais plutôt *des métadonnées* qui constituent le résultat d'une ou d'un certain nombre d'interprétations, d'analyses.

La figure 10 fournit un exemple d'une vue, d'un point de vue sur une donnée textuelle. L'objet est un segment vidéo qui illustre la formation d'un écosystème fluvial expliqué en voix off par un géographe. La vue sur cette donnée est exprimée sous forme d'un graphe d'individus (c'est-à-dire dont les objets d'analyse sont instanciés) qui d'une part, résume les caractéristiques d'expression du segment (du plan) et d'autre part, contient une description-interprétation du discours en voix off de l'expert. La figure 10 nous montre en effet une capture d'écran d'un environnement scientifique et technique développé entre 2000 et 2003 par l'INA sous la direction de Bruno Bachimont et plusieurs partenaires, dont l'équipe ESCoM-AAR[15] dans le cadre du projet de recherche OPALES[16].

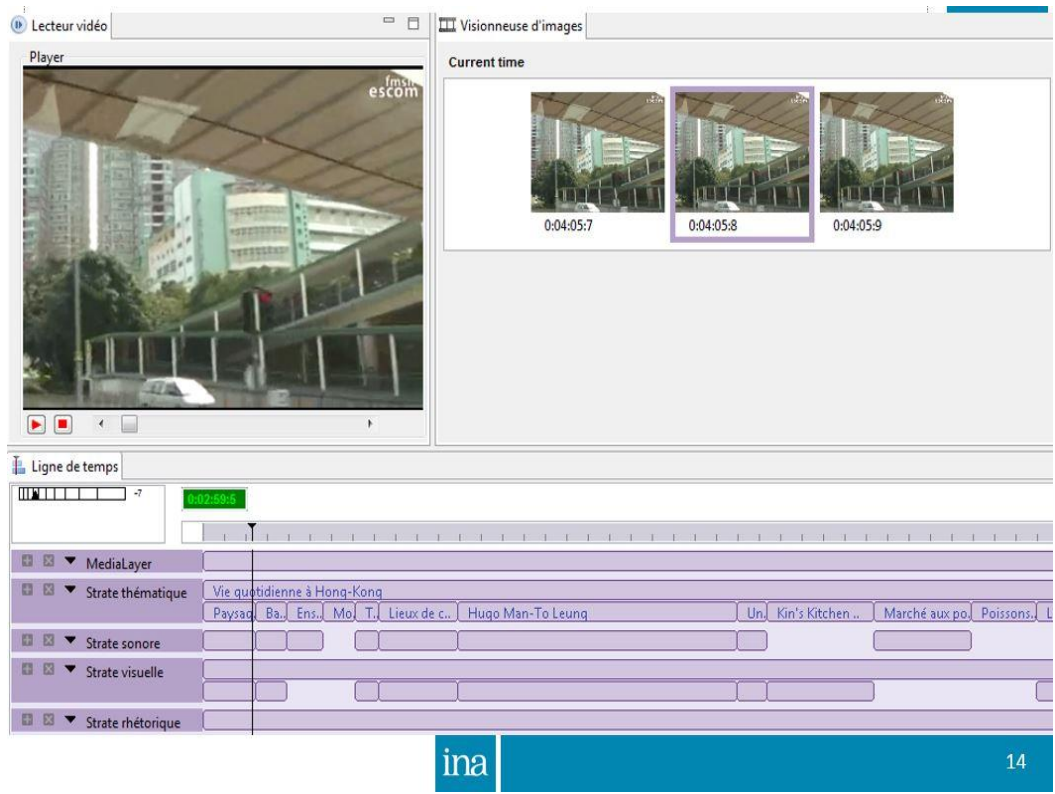

---

[15] Pour plus d'informations : https://halshs.archives-ouvertes.fr/ARC/hal-01330114v1
[16] Pour plus d'informations, cf. P. Stockinger: Digital audiovisual archives in humanities. Problems and challenges. Chania 2004 (en ligne: https://hal.archives-ouvertes.fr/hal-00130295)

(**Figure 11** : Interface « stand alone » OKAPI (2017), réalisée par Abdelkrim Beloued et Steffen Lalande, INA)

Les figures 11 et 12 montrent des exemples de visualisation de métadonnées à l'aide de la plateforme OKAPI : la première est en mode *stand alone* ; la seconde en mode *service web*. Un des ajouts les plus importants dont témoignent les deux figures 11 et 12 est celui du modèle de strates et de segments pour traiter l'objet « texte ». Modèle qui rend opérationnel l'image de Greimas du texte comme une « pâte feuilletée ».

Les figures 11 et 12 sont des visualisations du point de vue de l'analyste de données. Elles montrent la donnée elle-même (une vidéo) entourée par tout un ensemble de critères provenant du modèle conceptuel du domaine et que l'analyste peut utiliser pour expliciter et structurer la donnée en question.

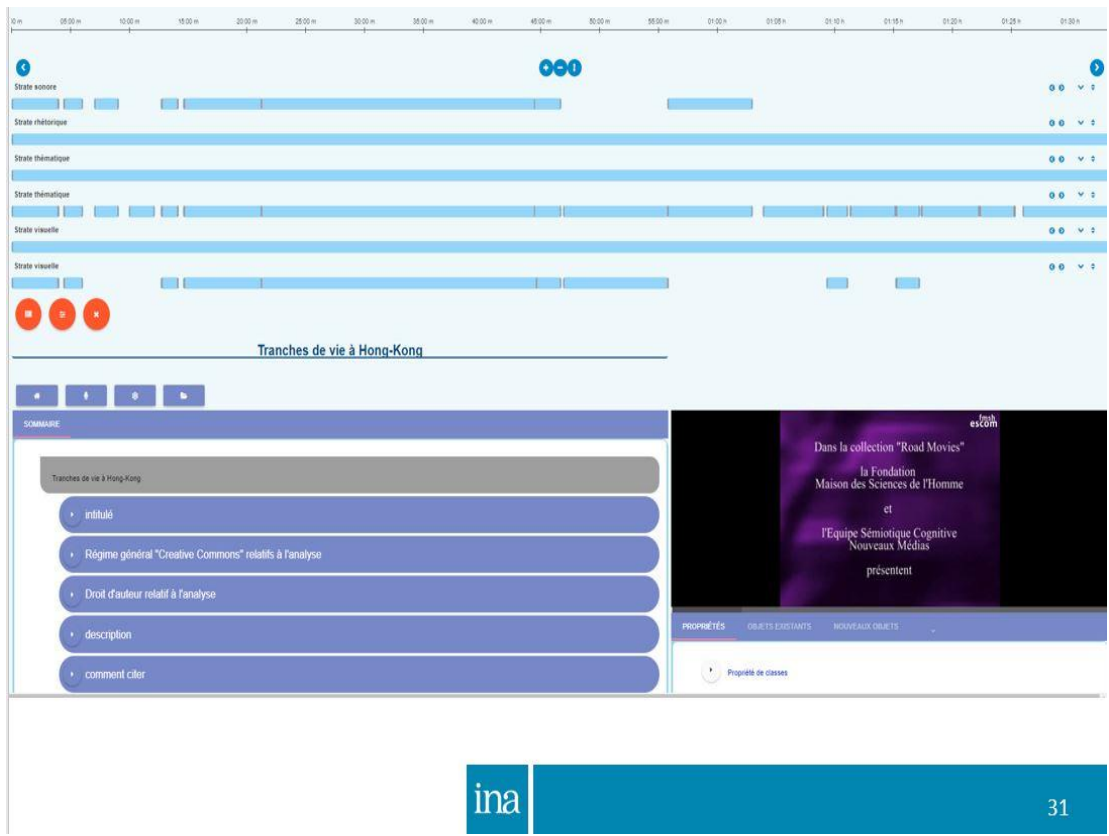

**Figure 12** : Interface web OKAPI (2018), réalisée par Abdelkrim Beloued et Steffen Lalande, INA)

Les figures 13 et 14, en revanche, montrent deux visualisations possibles des métadonnées du point de vue de l'utilisateur. La figure 13 montre une interface textuelle simple de visualisation de métadonnées pour accéder aux corpus de segments par strates, par thèmes, par sujets, par géolocalisation, via une timeline, selon la notoriété d'une ressource…

Enfin, la figure 14 montre la visualisation d'un segment de la strate thématique d'une vidéo et des métadonnées sous forme d'un graphe…

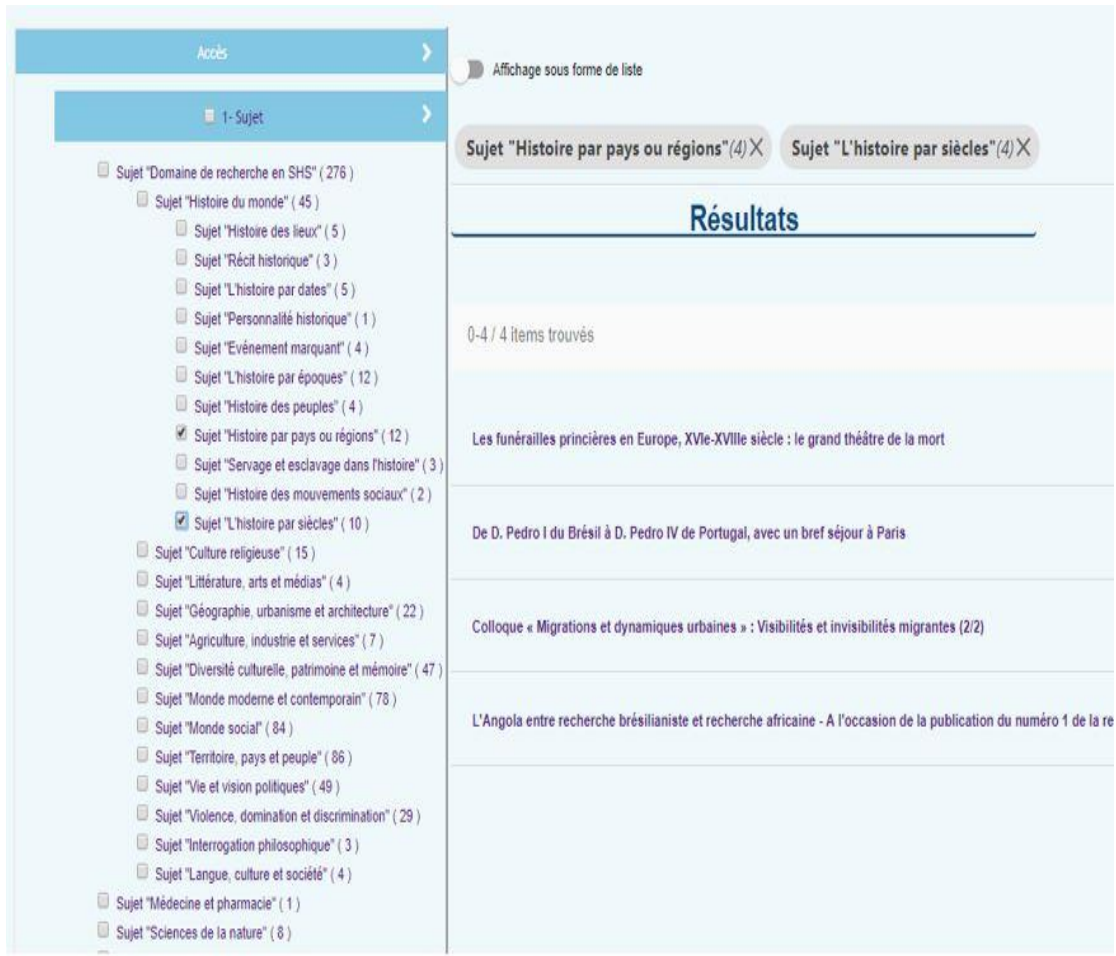

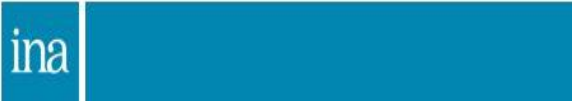

(**Figure 13** : Interface web OKAPI (2018), réalisée par Abdelkrim Beloued et Steffen Lalande, INA)

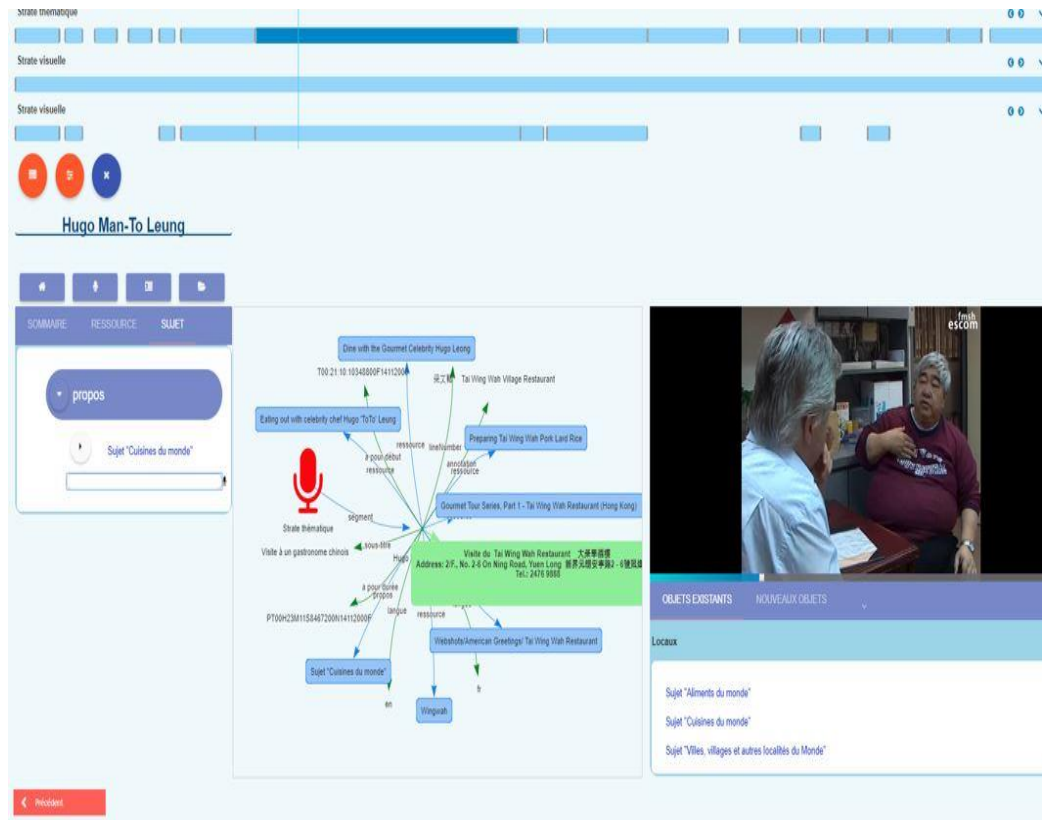

(**Figure 14** : Interface web OKAPI (2018), réalisée par Abdelkrim Beloued et Steffen Lalande, INA)

# 7) De la ré-ingénierie documentaire à l'acteur « texte » en incluant le visual storytelling

Un quatrième sens de visualisation de données que je voudrais encore introduire ici, c'est celui de la visualisation non pas (ou non pas exclusivement) des métadonnées, mais plutôt des *données textuelles* lato sensu *elles-mêmes*. Le texte lui-même se présente sous forme d'un certain **« habillage »** (verbal, graphique, iconique, acoustique, etc.) qui prend en charge l'**expression** du contenu, du message qu'il articule.

Or une activité importante aujourd'hui est celle de procéder au changement intentionnel de l'*habillage originel* d'un texte afin d'en faire une ressource appropriée dans le cadre d'un projet particulier donné - projet de communication, projet pédagogique, projet de médiation.

Alors, ce changement de l'habillage du texte est souvent accompagné de changement plus ou moins significatifs (c'est-à-dire *catégoriels*) du contenu du texte d'origine. Ici, la technique du design de l'information prend la forme de ce qu'on appelle (en anglais) *document repurposing*, en français *ré-ingénierie documentaire*, *réécriture*, *ré-éditorialisation* ou encore *republication* : avec des données préexistantes, on produit un *nouveau texte*, une *nouvelle prestation*, une *nouvelle donnée*. Citons ici, par exemple, la suite logicielle *Scenari*[17] dédiée à la republication de données documentaires ainsi que le portail de la communauté Scenari.org[18] qui s'intéresse plus particulièrement à la réutilisation de données documentaires en différents formats.

Changer l'habillage originel d'un texte va systématiquement de pair avec un changement intentionnel du contenu du texte d'origine. C'est typiquement le cas du montage, compris comme une technique culturelle d'appropriation et de réutilisation de données textuelles produites en général par un ou plusieurs auteurs. Nous pensons, par exemple, à cette catégorie particulière de productions audiovisuelles qui circulent sur les médias sociaux et dont les auteurs sont souvent des anonymes. Ces derniers – d'une manière analogue aux *braconniers* de M. de Certeau – prélèvent des segments audiovisuels, visuels, acoustiques, etc. qui circulent déjà sur les réseaux sociaux pour en faire de *nouvelles réalisations* audiovisuelles. Pour donner un exemple, nous nous sommes intéressés depuis déjà plusieurs années à cette quantité assez colossale de vidéos sur les réseaux sociaux qui thématisent la figure du *migrant* et, plus particulièrement, celle de l'*immigré* (en Europe, en Australie ou encore aux États-Unis et au Canada). Parmi ces milliers de vidéos, un nombre assez significatif est réalisé par des anonymes ou par des individus qui animent des réseaux, des communautés d'adhérents plus ou moins importantes. Très souvent, ces vidéos ne sont pas des réalisations sui generis (dont les producteurs restent les « grands médias » classiques, les ONG, les OIG…). Il s'agit ici plutôt de remontages (« remix », en anglais) de bouts de vidéos accompagnés d'une musique d'accompagnement et parfois des explications en voix in ou en voix off. L'ensemble de ce montage a, bien évidemment, comme objectif de convaincre le public visé ou réel de la pertinence, de la véracité du propos des auteurs de ces montages. Or, on rencontre ici en effet, une assez importante catégorie de remontages qui dénonce l'envahissement de l'Europe (de l'Australie, des États-Unis…) et de la civilisation judéo-chrétienne par les immigrés arabes, musulmans, mâles, se déplaçant en hordes avec l'intention de remplacer la civilisation occidentale. Ces productions qu'on qualifie d'une manière un peu simpliste comme appartenant à la « fachosphère » procèdent par la sélection thématiquement et visuellement pertinente de séquences souvent très courtes qui mettent en scène des actes d'agression, des hurlements de masses d'immigrés, des scènes d'affrontement entre manifestants et policiers, etc.

Le résultat est un contenu particulièrement violent qui se nourrit de réalisations dont les auteurs sont les médias, les ONG et autres acteurs de l'opinion publique de la société contemporaine. Cela dit, le *re-design conceptuel* et le *relookage visuel (lato sensu)* d'un contenu ne sont pas l'exclusivité de « petits fachos ». Cette façon de manipuler intentionnellement contenu et expression, on la trouve partout également parmi celles et ceux qui s'engagent, au contraire, pour une vision humaniste, emphatique… de l'immigration. Autrement dit, la technique du *repurposing* s'appuie sur des pratiques rhétoriques fort anciennes et traditionnelles, transversales à toute activité de communication, à tout écosystème de communication.

L'intérêt de discuter cette perspective très particulière de la visualisation d'une information réside dans le fait que nous commençons aujourd'hui à disposer de connaissance et d'outils indispensables pour expérimenter avec celle-ci, pour en faire une sorte de technologie qui à la fois « pompe » dans nos connaissances et traditions, mais qui l'enrichissent aussi pour en faire de dispositifs « intelligents » de communication, de partage et d'exploitation de connaissance.

---

[17] https://scenari.kelis.fr/
[18] https://scenari.org/co/home.html

Ainsi, parmi les tendances de la *visualisation de données* aujourd'hui très prisées, nous trouvons le *visual story telling*, le *visual data story telling* ou encore les *mash up*. Généralement parlant, ces techniques s'appuient sur un fonds d'images et de vidéos pour faire découvrir un domaine de connaissance. Prenons de nouveau notre exemple étudié dans le projet Memomines, c'est-à-dire celui du monde des mines. Pour réaliser un *visual storytelling* sur la mise en scène de la visite d'une mine comprise comme un lieu de travail, il nous faudra :

1. un modèle conceptuel du monde des mines ;
2. un *contenu approprié* sous forme de corpus de segments vidéo ;
3. un *scénario narratif* de narration selon lequel le contenu sera mis en récit, relaté ;
4. un *modèle de visualisation* (lato sensu) du récit de la visite d'une mine et des éléments du corpus d'analyse qui peuvent servir de documentation de cette visite.

Pour le scénario narratif, il est typiquement organisé sous forme de *parcours narratifs* de différents genres. Le plus simple parcours est celui de la succession linéaire d'étapes de la visite d'une mine, avec laquelle cette succession est prédéterminée. Mais, on peut également s'imaginer des structures plus complexes qui ressemblent, par exemple, à des arbres de décision avec un ou plusieurs étapes finales. Dans une étape donnée, plusieurs chemins pour s'en aller peuvent être possibles : le fait d'emprunter certains chemins peut être soumis à certaines conditions (comme, dans un cours ou dans un « jeu sérieux ») ;, etc. (voir Stockinger et al. 1992).

L'important ici est que le parcours narratif est de nouveau décrit comme un graphe dont les sommets sont les étapes et les arêtes les chemins entre étapes. Le sommet représentant une étape peut se déployer, s'expanser en un graphe représentant la structure d'une étape, etc.

Le fait vraiment important ici est celui d'avoir une bonne théorie empirique – un bon modèle – de l'objet *narration,* avec d'une manière élémentaire une typologie minimale des étapes narratives et des chemins narratifs menant d'une étape à une autre. Une telle ontologie élémentaire de la narration pourra ensuite être complexifiée, enrichie.

Les données audiovisuelles qui font partie du corpus analysé seront appariées (« matchées ») avec les étapes d'un parcours. Autrement dit, elles figureront dans les étapes pour lesquelles elles apportent les informations pertinentes, appropriées. Pour faire cela, il faut que les différents types d'étapes intègrent dans leur définition plus particulièrement des spécifications thématiques, narratives, discursifs, rhétoriques, visuelles et sonores.

Enfin, le modèle de visualisation (lato sensu, c'est-à-dire pas obligatoirement réduit à la seule modalité perceptive de la vision) doit définir la *mise en scène* de ce *storytelling* s'appuyant sur des données structurées et liées. Le *visual (data) storytelling* fait en effet, appel à une bibliothèque de *widgets sémantiques*, c'est-à-dire de composants de site définis dans une *ontologie de narration* et de l'*éditorialisation* de la narration des données audiovisuelles.

Sur la base de ces trois points – *scénario narratif ; contenu approprié ; visualisation des données/métadonnées* – on voit donc se développer des applications « auteur » qui réutilisent des données préexistantes, qui leur donnent une *seconde vie*, qui les utilisent comme données pour relater toutes sortes d'histoire. Ces applications peuvent correspondre à une *nouvelle œuvre figée* (à une nouvelle vidéo augmentée). Mais, elles peuvent être aussi des applications *ouvertes*, se renouvelant par de nouvelles données et par des structures narratives personnalisables. Notons que ces applications « auteur » simultanément dynamiques (ouvertes aux nouvelles données) et personnalisables répondent aux nouvelles possibilités et pratiques de lecture et d'appropriation de contenus numériques telles que les pratiques d'appropriation collaborative et participative, les agrégateurs (sémantiques) de données textuelles, les newsmaps, les infographies interactives, les (ré)écritures à temps réel, les « news river » ou encore l'appropriation de l'information ambiante via une interface ubiquitaire.

La problématique générale qui nous intéresse ici plus particulièrement est celle de *faire parler* les fonds de données des archives en agissant pour qu'elles se comportent un peu comme des *acteurs* qui répondent aux demandes, aux besoins des utilisateurs des archives qu'ils « savent » agir seul ou d'une manière collective pour résoudre un problème ou satisfaire une demande, etc. Métaphoriquement parlant, nous aimerions les voir comme des *objets enchantés*, des *enchanted objects* pour parler avec David Rose (voir son ouvrage populaire *Enchanted Objects: Design, Human Desire, and the Internet of Things*, New York, Scribner 2014).

Cela veut dire qu'un texte, qu'une donnée textuelle (lato sensu) devrait être, métaphoriquement parlant, « conscient » de sa structure sémantique, des différentes manières et des stratégies qu'il utilise pour exprimer, mettre en scène son contenu et de l'environnement sémiotique dans lequel il agit. Ainsi, en étant « conscient » de son *profil* et de sa *place* dans un environnement donné, il pourrait interagir avec d'autres textes ou avec un acteur humain ou automate pour contribuer simultanément à la production, à la génération de toutes sortes de récits, et d'environnements signifiants, d'allosphères inclusifs qui font échos aux souhaits, besoins, intérêts, désirs, imaginations d'un acteur. Bref, l'image de référence ici est celle des textes qui sont simultanément des ressources cognitives et des agents regroupés en communautés, en sociétés à thème, axiologiques et contractuelles (en référence aux recherches en intelligence artificielle distribuée et sur les sociétés artificielles).

À notre avis, c'est une des perspectives les plus excitantes d'une ingénierie sémiotique théorique et appliquée des ressources textuelles dans une perspective du web sémantique et, au-delà, d'une économie du sens (« meaning economy », en anglais) qui émerge progressivement.

# Bibliographie